\documentclass[runningheads]{llncs}
\usepackage{amssymb}
\usepackage[T1]{fontenc}
\usepackage{graphicx}
\usepackage{amsmath}
\usepackage[colorlinks=true, linkcolor=blue, urlcolor=blue, citecolor=blue]{hyperref}
\usepackage{orcidlink}
\usepackage{color}

\usepackage{booktabs}
\usepackage{float}
\usepackage{adjustbox}
\usepackage{comment}
\usepackage{marvosym}

\begin{document}
\title{GaussianFocus: Constrained Attention Focus for 3D Gaussian Splatting}
%
%
\author{Zexu Huang\textsuperscript{({\Letter})}\orcidlink{0009-0006-6251-9694} \and
Min Xu\orcidlink{0000-0001-9581-8849} \and
Stuart Perry\orcidlink{0000-0002-2794-3178}}
\authorrunning{Z. Huang et al.}
%
\institute{University of Technology Sydney, Ultimo NSW 2007, Australia
\email{Zexu.Huang@student.uts.edu.au, \{Min.Xu,Stuart.Perry\}@uts.edu.au}}
\maketitle              
\begin{abstract}
Recent developments in 3D reconstruction and neural rendering have significantly propelled the capabilities of photo-realistic 3D scene rendering across various academic and industrial fields. The 3D Gaussian Splatting technique, alongside its derivatives, integrates the advantages of primitive-based and volumetric representations to deliver top-tier rendering quality and efficiency. Despite these advancements, the method tends to generate excessive redundant noisy Gaussians overfitted to every training view, which degrades the rendering quality. Additionally, while 3D Gaussian Splatting excels in small-scale and object-centric scenes, its application to larger scenes is hindered by constraints such as limited video memory, excessive optimization duration, and variable appearance across views. To address these challenges, we introduce GaussianFocus, an innovative approach that incorporates a patch attention algorithm to refine rendering quality and implements a Gaussian constraints strategy to minimize redundancy. Moreover, we propose a subdivision reconstruction strategy for large-scale scenes, dividing them into smaller, manageable blocks for individual training. Our results indicate that GaussianFocus significantly reduces unnecessary Gaussians and enhances rendering quality, surpassing existing State-of-The-Art (SoTA) methods. Furthermore, we demonstrate the capability of our approach to effectively manage and render large scenes, such as urban environments, whilst maintaining high fidelity in the visual output. Please visit \href{https://github.com/HZXu-526/GaussianFocus}{https://github.com/HZXu-526/GaussianFocus} for code.

\end{abstract}
\section{Introduction}
Novel View Synthesis (NVS) is fundamental for modern computer graphics and vision, extending to virtual reality, autonomous driving, and robotics. Primitive-based models such as meshes and point clouds~\cite{lassner2021pulsar,munkberg2022extracting,yifan2019differentiable}, optimized for GPU rasterization, deliver fast but often lower-quality images with discontinuities. The introduction of Neural Radiance Fields (NeRF)~\cite{mildenhall2021nerf} marked a significant advancement, employing a multi-layer perceptron (MLP) to achieve high-quality, geometrically consistent renderings of novel viewpoints. However, NeRF's reliance on time-consuming stochastic sampling can lead to reduced performance and potential noise issues.

Recent advancements in 3D Gaussian Splatting (3DGS)~\cite{kerbl20233d} have significantly enhanced rendering quality and speed. This technique refines a set of 3D Gaussians, initialized using Structure from Motion (SfM)~\cite{snavely2006photo}, to model scenes with inherent volumetric continuity. This facilitates fast rasterization by projecting them onto 2D planes. However, 3DGS often produces artifacts when camera viewpoints deviate from the training set and lack detail during zooming. To address these issues, newer models~\cite{yu2024mip,lu2024scaffold} employ a 3D smoothing filter to regularize the frequency distribution and utilize anchor points to initialize 3D Gaussians, thereby enhancing visual accuracy and applicability in diverse scenarios. Despite these advances, 3DGS-based models still tend to produce oversized Gaussian spheres that ignore scene structure, leading to redundancy and scalability issues in complex environments. Additionally, these models struggle with detail reconstruction, particularly at edges and high-frequency areas. This often leads to suboptimal rendering quality. Moreover, reconstructing large-scale scenes like towns or cities represents a significant challenge due to GPU memory constraints and computational demands. To mitigate these problems, models often reduce the number of training inputs randomly, which compromises reconstruction quality and results in incomplete outcomes.

\begin{figure}[t]
    \centering
    \includegraphics[width=0.9\textwidth]{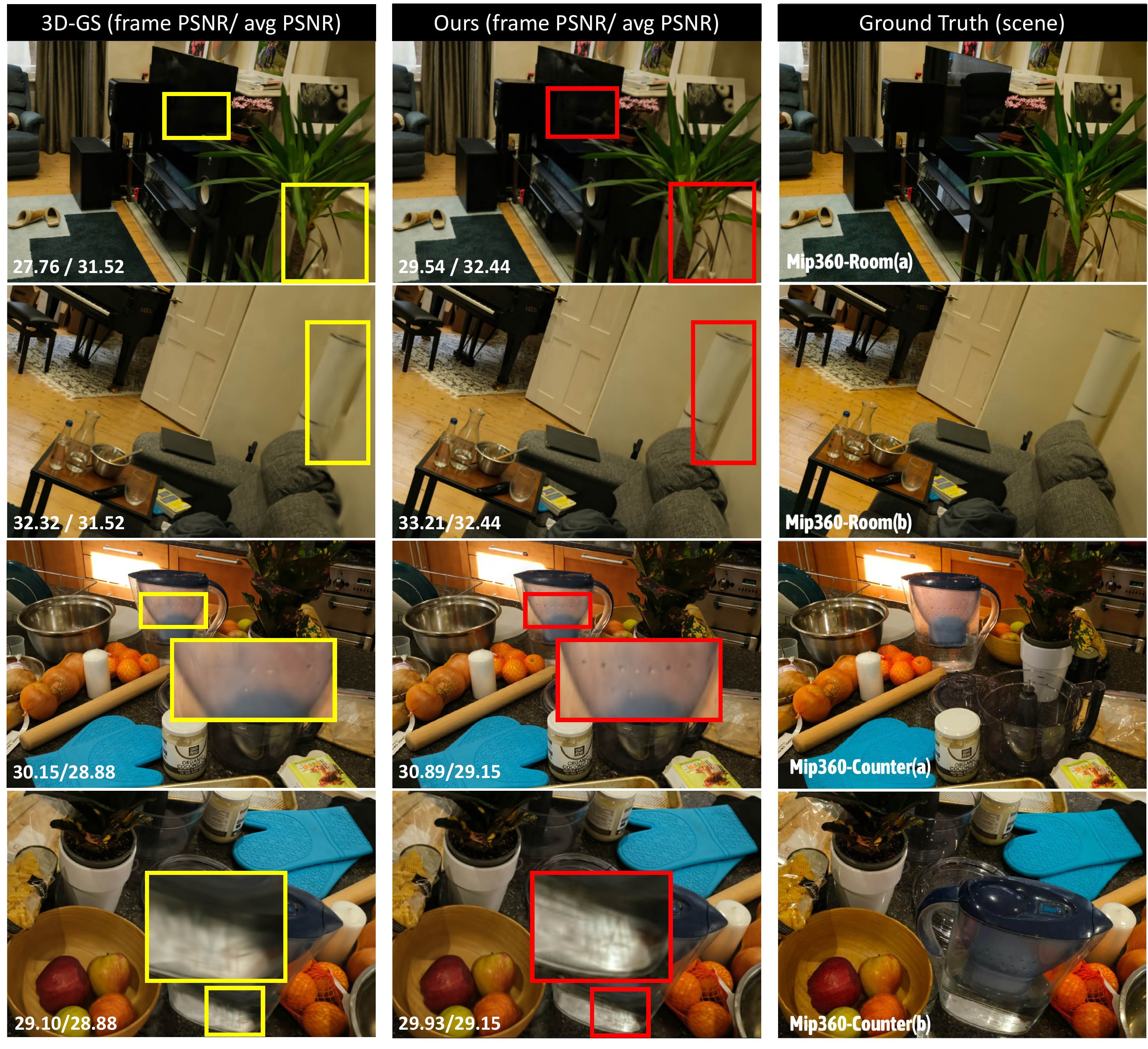}
    \caption{\textbf{GaussianFocus.} As illustrated by the red and yellow boxes in the images, our method consistently surpasses the 3DGS model in various scenes, showing distinct advantages in challenging environments characterized by slender geometries, intricate details, and lighting effects.}
    \label{fig:compare_3dgs}
\end{figure}

To address quality issues in 3D Gaussian Splatting (3DGS), we introduce GaussianFocus, a framework designed for enhanced fidelity in both general and large-scale scene reconstructions. GaussianFocus employs a patch attention algorithm and Sobel operators to refine edge details and spatial frequency during training, thereby improving scene fidelity. We also apply constraints on the size of Gaussian spheres during initialization and training phases, which refines texture details and diminishes the occurrence of ``air walls''. These ``air walls'' are spurious barriers or noise in 3D reconstructions, typically resulting from oversized Gaussian spheres that disrupt visual coherence. For reconstructing extensive scenes, our method uses bounding boxes to divide each scene along the XYZ axes into manageable blocks. Each block is independently processed in our 3D reconstruction pipeline, ensuring precise attention to its specific features. After processing, these blocks are seamlessly recombined, producing a coherent and detailed large-scale reconstruction.

Through rigorous experiments, our GaussianFocus model has outperformed traditional 3DGS models~\cite{kerbl20233d}, as evidenced in Fig.~\ref{fig:compare_3dgs}. It notably reduces artifacts associated with oversized Gaussian spheres, thereby enhancing the quality of 3D reconstructions. Our subdivision strategy for large-scale scenes considerably lowers GPU computational demands, allowing for the use of all input data and maintaining superior reconstruction quality. This represents a significant improvement over previous approaches~\cite{kerbl20233d,yu2024mip,lu2024scaffold,guedon2024sugar}, which often required sub-sampling of input data to manage computational loads. GaussianFocus thus significantly improves the realism and quality of 3D reconstructions.


In summary, the contributions of this work are as follows:

\begin{enumerate}
    \item We propose a 3DGS-based patch attention algorithm with novel edge and frequency losses designed to enhance details and reduce spatial frequency artifacts within scene reconstructions. This improves the detailing quality and intricacy of the rendered scenes.

    \item We impose constraints on overly large Gaussian spheres to mitigate the occurrence of ``air walls'', thus improving the scene reconstruction fidelity and enhancing the granularity of the resulting models. Moreover, these constraints allow the achievement of superior reconstruction results with fewer training iterations.

    \item For large-scale scene reconstruction, our approach involves subdividing the scene for reconstruction and subsequent recombination. This method addresses the challenge posed by existing 3DGS-based models that fail to directly reconstruct extensive scenes, thereby enhancing the scalability and applicability of our reconstruction framework.
\end{enumerate}

\section{Related Work}
\noindent \textbf{Volumetric Rendering methods}\ \ \ \
Volumetric approaches represent scenes as continuous functions specifying volume, density, and color characteristics. The introduction of Neural Radiance Fields (NeRF)~\cite{mildenhall2021nerf} was a pivotal advancement, using a multilayer perceptron (MLP) to parameterize these functions and create photorealistic images that vary with viewer perspective through volumetric ray tracing. However, the computational and memory demands of vanilla NeRF limit its practical application. Subsequent research has improved NeRF's efficiency and scalability through discretized or sparse volumetric frameworks like voxel grids and hash tables~\cite{chen2022tensorf,sun2022direct,muller2022instant}, which hold learnable features acting as positional encodings for 3D coordinates. Additionally, some methods employ hierarchical sampling techniques~\cite{barron2022mip,reiser2021kilonerf,yu2021plenoctrees} and utilize low-rank approximations~\cite{chen2022tensorf}. Despite these enhancements, the dependence on volumetric ray marching continues, which leads to compatibility challenges with traditional graphics equipment and systems primarily engineered for polygonal rendering.

\vspace{1em}

\noindent \textbf{Large-scale Scene Reconstruction}\ \ \ \
Concurrently, a plethora of adaptations have surfaced as Neural Radiance Fields (NeRF)~\cite{mildenhall2021nerf} gain prominence for generating photorealistic perspectives in contemporary visual synthesis. These aim to increase reconstruction quality~\cite{barron2021mip,barron2022mip,barron2023zip,wang2021neus,wang2023neus2}, accelerate rendering~\cite{chen2022tensorf,muller2022instant}, and extend capabilities to dynamic scenarios~\cite{weng2022humannerf,huang2024efficient}. Among these, several methods~\cite{tancik2022block,turki2022mega} have scaled NeRF to accommodate expansive scenes. However, they often render slowly and lack finer details. The newly introduced 3D Gaussian Splatting (3DGS)~\cite{kerbl20233d} enables explicit, high-definition 3D representations with real-time rendering capabilities. However, applying traditional 3DGS methods~\cite{kerbl20233d,yu2024mip,lu2024scaffold,huang2025structgs} to large scenes, such as urban or scenic landscapes, demands significant memory and graphics resources for initial scene processing and Gaussian sphere creation. Several works~\cite{lin2024vastgaussian,kerbl2024hierarchical} have scaled 3DGS to large-scale scenes, but their methods fail to adapt to large changes in viewpoint. Previously, methods~\cite{kerbl20233d,yu2024mip,lu2024scaffold} could not directly process large-scale scenes and instead relied on selecting a reduced subset of input images to perform pre-processing using COLMAP~\cite{schonberger2016structure}, which applies Structure-from-Motion (SfM) and Multi-View Stereo (MVS) to estimate camera poses and sparse point clouds. However, this strategy often led to incomplete or fragmented scene reconstructions due to suboptimal scene coverage and loss of important visual information. Moreover, using incomplete image sets often resulted in poor COLMAP~\cite{schonberger2016structure} outcomes, failing to cover all necessary viewpoints for comprehensive scene reconstruction. Our GaussianFocus successfully overcomes the limitations of the 3DGS-based methods~\cite{kerbl20233d,yu2024mip,lu2024scaffold} in training large-scale scenes through the introduction of methods that efficiently subdivide, optimize, and integrate these scenes.
\section{Methodology}
\label{sec:method}

\begin{figure}[t]
    \centering
    \includegraphics[width=\textwidth]{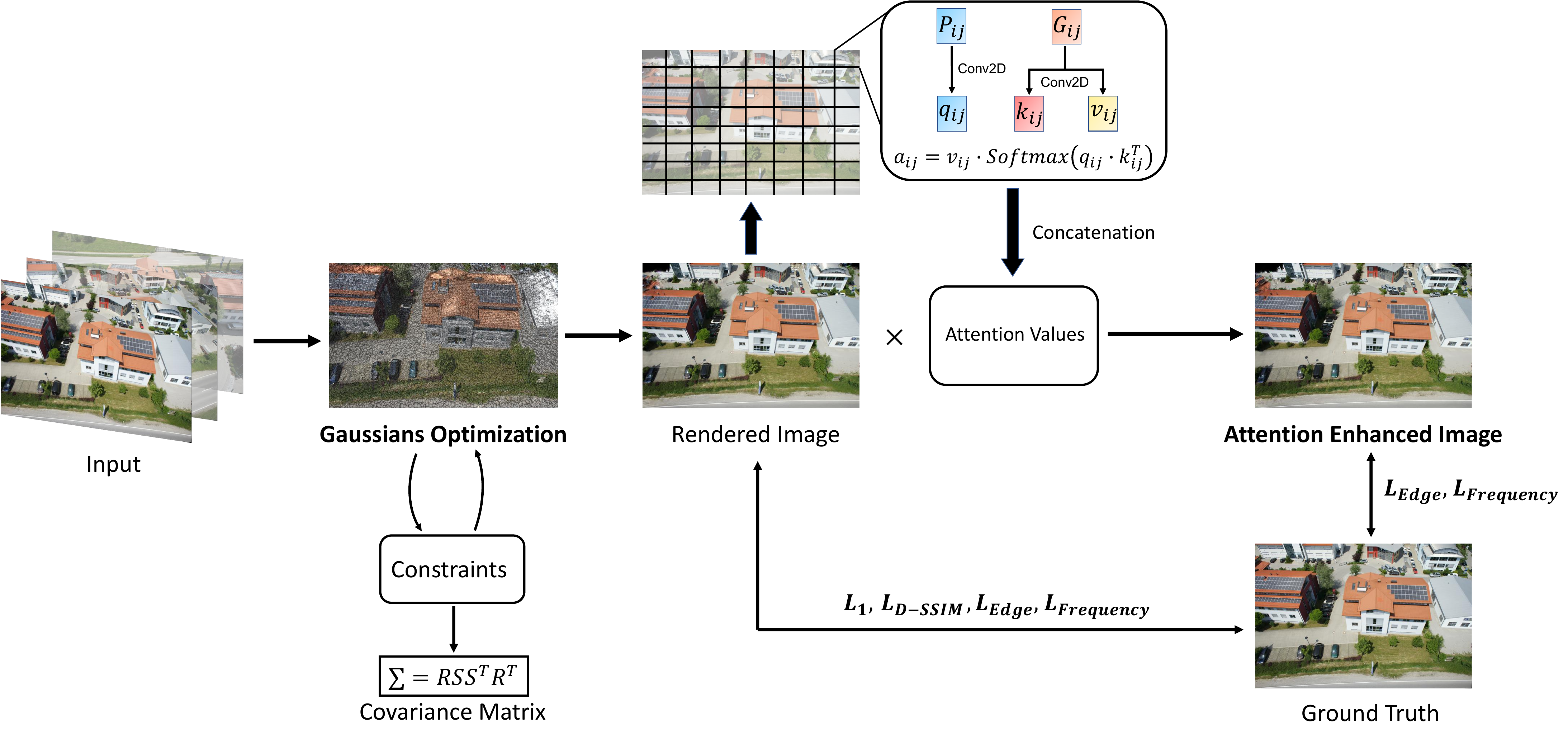}
    \caption{\textbf{Overview of GaussianFocus:} Our model monitors the size of Gaussian spheres during initialization and training. \textbf{Constraints} are applied to the scaling matrix $S$ within the covariance matrix to prevent the excessive growth of the Gaussian spheres. Subsequently, the rendered image is divided into 64 parts. Each part independently calculates its attention values, which are then concatenated to form a comprehensive attention map. This map is multiplied back onto the original rendered image to produce an \textbf{attention-enhanced image}. Finally, this enhanced image and the original rendered image undergo multiple loss calculations against the ground truth. These include reconstruction ($L_1$), structural similarity ($L_{D-SSIM}$), edge ($L_{Edge}$), and frequency ($L_{Frequency}$) losses.}
    \label{fig:model_structure}
\end{figure}

The traditional 3DGS~\cite{kerbl20233d} and its variants~\cite{yu2024mip,guedon2024sugar,lu2024scaffold} employ Gaussian optimization to reconstruct scenes, often failing to accurately represent actual scene structures and struggling with oversized Gaussians that blur scenes and lead to information loss. Limited GPU memory and extended optimization times further hinder their ability to reconstruct large scenes. Our enhanced framework, detailed in Fig.~\ref{fig:model_structure}, addresses these limitations by imposing constraints on the size and quantity of 3D Gaussian spheres, reducing redundancy and improving robustness against varying viewing conditions. We incorporate attention mechanisms and a combination of edge and frequency loss to refine reconstruction quality.
\subsection{3D Gaussian-Based Patch Attention Enhancement}
\label{sec:attention}
Given the significant computational demands, it is impractical to directly compute attention values for the entire rendered image due to the extensive data processing involved. Instead, both model-rendered image $P_i$ and the Ground Truth images $G_i$ are segmented into 8x8 regions to manage computational complexity effectively. For each segment of $P_i$, a query vector $q_{ij}$ is extracted using a 2D convolutional layer, which is designed to capture detailed features and spatial relationships within the segment. Correspondingly, the key $k_{ij}$ and value $v_{ij}$ for each segment $j$ of $G_i$ are derived through similar 2D convolutional layers. These steps ensure that the essential components for the multi-head attention mechanism—queries, keys, and values (QKV)—are accurately assembled based on localized image features. The attention weights $w_{ij}$ for each segment can be calculated using the following equation:
\begin{equation}
    w_{ij} = \text{Softmax}(\alpha_{ij}), \quad \alpha_{ij} = q_{ij} \cdot k_{ij}^T,
\end{equation}
where $\alpha_{ij}$ represents the unnormalized attention scores, which are computed as the dot product of the query and the transposed key. This product measures the compatibility between different parts of the image, facilitating a focused synthesis of features. The attention map for each segment $a_{ij}$ is generated by applying the weighted sum of the values using the attention weights:
\begin{equation}
    a_{ij} = v_{ij} \cdot w_{ij},
\end{equation}
where $w_{ij}$ scales the value $v_{ij}$ according to the relevance of each segment's features, thereby producing a segment-specific attention map that highlights pertinent features. Concatenating these individual attention maps yields a comprehensive attention map $A_i$ for the image, which can be represented by:
\begin{equation}
    A_i = \bigoplus_{j} a_{ij}
\end{equation}
where the sum over $j$ aggregates and assembles the contributions of all segments into a unified attention map for the entire image. This comprehensive attention map $A_i$ is then used to produce an attention-enhanced image $P_{i}^{'}$ by element-wise multiplying it with the rendered image $P_i$:
\begin{equation}
    P_{i}^{'} = P_i \otimes A_i,
\end{equation}
which enhances the original image by amplifying features that are deemed significant based on the attention mechanism. To further enhance the reconstruction accuracy, we compute edge loss $L_{Edge}$ and frequency loss $L_{Frequency}$ for this enhanced image in conjunction with the ground truth image. These losses are calculated alongside the standard loss comparisons between the original rendered image and the ground truth image. They will be discussed in Section~\ref{sec:losses}.
\subsection{Gaussian Sphere Constraints}
\label{sec:gaussian constraints}
During the initialization of Gaussian spheres, we impose constraints on the scaling matrix $S$ to control the covariance matrix's influence, essential for accurately modelling spatial relationships in the scene. The adjustment of $S$ is dictated by the density of the initial point cloud data: for denser point clouds, we set a lower initial scaling value to reduce overlaps and redundancy, while for sparser distributions, we increase it to ensure sufficient scene coverage. We used the distance to determine the density for the point clouds. This careful calibration of scaling factors helps maintain an optimal balance between preserving detail and enhancing computational efficiency. The scaling matrix constraint is defined as follows:
\begin{equation}
S_i = S_i \cdot \alpha, \quad \text{if } S_i > \tau, \\
\end{equation}
where \( S_i \) denotes the scales in the scaling matrix of the Gaussians. The \(\tau\) serves as a threshold scale and \(\alpha\) is a modulating factor, both of them adjusted experimentally. In our experiment, we set \(\tau\) = 0.3 and \(\alpha\) = 0.2. The adaptive scaling in our model not only mitigates computational load but also aligns with the varying densities of real-world data. Enhancing the traditional ``split and clone'' strategy of the 3DGS~\cite{kerbl20233d} model, we integrate a filtering mechanism to manage excessively large Gaussians during training. This involves implementing a selection criterion to identify large Gaussians post-splitting, followed by a strategic reduction in their scale. Additionally, we employ a selective splitting strategy for older Gaussians that have remained in the model over extended periods. This technique is based on both the age and the operational efficiency of the Gaussian in terms of scene representation:
\begin{equation}
\text{Selective Split} \; (S_{\gamma}), \quad \text{if } S_{\gamma} > \Omega
\end{equation}
where \( S_{\gamma} \) denotes the scales of the scaling matrix of aged Gaussians and \(\Omega\) is the scale threshold set to identify old Gaussians that require reevaluation. We set \(\Omega\) = 0.3 in our experiment. This strategy reduces the unnecessary split in training. These strategies ensure that our method maintains a balanced approach to managing the size and number of Gaussians within the 3DGS framework.
\subsection{Subdivision-Based Reconstruction of Large Scenes}
\label{sec:segment gaussian}
\begin{figure*}
    \centering
    \includegraphics[width=1\textwidth]{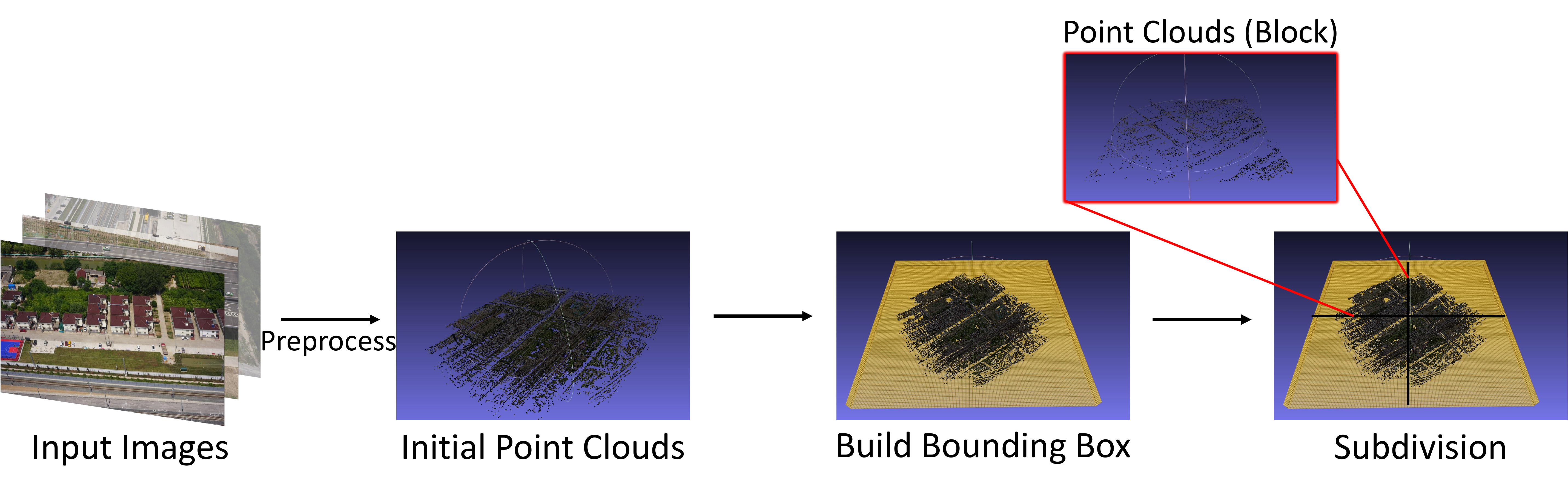}
    \caption{\textbf{Subdivision-Based Reconstruction of Large Scenes Procedure.} Our method divides large scenes into blocks for reconstruction.}
    \label{fig:segment_method}
\end{figure*}

In response to 3DGS challenges~\cite{kerbl20233d,yu2024mip,lu2024scaffold}, our method initiates a preprocessing step to acquire initial points from Structure-from-Motion (SfM)~\cite{snavely2006photo} of the large scene. As shown in Fig.~\ref{fig:segment_method}, a three-dimensional bounding box is then constructed to encompass initial point clouds. We divide this bounding box along its \(xyz\) axes into \(n \times n \times n\) blocks, where each block is defined to contain its respective subset of point clouds:
{\footnotesize
\begin{equation}
B_{ijk} = \left\{ pc \in \text{Point Cloud} : (x_i \leq pc_x < x_{i+1}) \wedge (y_j \leq pc_y < y_{j+1}) \wedge (z_k \leq pc_z < z_{k+1}) \right\},
\end{equation}
}
where \(pc\) represents a point in the point cloud and \(x_i, y_j, z_k\) denote the boundaries of block \(B_{ijk}\). We have integrated a distance iteration algorithm to address the potential for sparse outlier points to skew the subdivision logic. This algorithm iterates through all points, identifying and discarding those that do not contribute meaningfully to the division process:
\begin{equation}
\text{Iterate } \forall pc \in \text{Point Cloud}: \text{if } \text{dist}(pc, \text{Center}_{ijk}) > \theta \text{ then discard } pc,
\end{equation}
where \(\text{dist}(\cdot)\) calculates the distance from the point to the block center, and \(\theta\) is a threshold value defining the maximum allowable distance for inclusion. Corresponding camera and Structure-from-Motion (SfM) points associated with each block are classified to assemble the essential initial inputs required for training. Each block undergoes independent training. The process concludes with the recombination of the divided scene's 3D files, thus completing the reconstruction of the entire large scene. This modular approach alleviates the computational and memory constraints typically linked with large-scale scene reconstruction. By employing this method, we efficiently manage large scene datasets and enhance the scalability of our reconstruction processes.
\subsection{Training Losses}
\label{sec:losses}
In our GaussianFocus model, following 3DGS, the loss function incorporates both L1 and D-SSIM terms. The L1 term measures absolute differences between predictions and targets, while D-SSIM enhances perceptual image and video quality. To improve the structural accuracy during training, we designed an edge loss term that leverages the Sobel operator to extract edge information effectively. This operator is applied to each channel of both the input and target images to compute their respective gradients in the \( x \) and \( y \) directions. The edge loss is then calculated as the average of the L1 loss of these gradients:
\begin{equation}
L_{\text{Edge}} = \frac{1}{2} \left( \text{L1}(\nabla_x p_i, \nabla_x \hat{p}_i) + \text{L1}(\nabla_y p_i, \nabla_y \hat{p}_i) \right),
\end{equation}
where \( \nabla_x \) and \( \nabla_y \) represent the gradient operator calculated using the Sobel filter, capturing edge information along the $x$ and $y$ directions. The variables $p_i$ and $\hat{p}_i$ represent the pixels of the ground truth image $G_i$ and the corresponding pixel in the rendered image $P_i$, respectively. Moreover, we introduce a frequency loss term to address the challenge of high-frequency detail loss. It approximates the frequency domain loss by employing gradient loss computations in the \( x \) and \( y \) directions for both the input and target images. This term is essential for preserving high-frequency details and is computed as:
\begin{equation}
L_{\text{Frequency}} = \frac{1}{2} \left( \text{L1}(G_x(p_i), G_x(\hat{p}_i)) + \text{L1}(G_y(p_i), G_y(\hat{p}_i)) \right),
\end{equation}
where $G_x$ and $G_y$ are the changes in pixel values along the horizontal and vertical axes. The overall loss function for the GaussianFocus model integrates these individual loss components into a weighted sum, optimizing the reconstruction quality across multiple dimensions:

\begin{equation}
\begin{split}
L_{\text{Total}} = \begin{cases}
    (1 - \lambda)L_{1}(p_i, \hat{p}_i) + \lambda L_{\text{D-SSIM}}(p_i, \hat{p}_i) + \\ \beta L_{\text{Edge}}(p_i, \hat{p}_i^{'}) + \eta L_{\text{Frequency}}(p_i, \hat{p}_i^{'}),\quad \text{every 50 iterations,}\\
    (1 - \lambda)L_{1}(p_i, \hat{p}_i) + \lambda L_{\text{D-SSIM}}(p_i, \hat{p}_i) + \\ \beta L_{\text{Edge}}(p_i, \hat{p}_i) + \eta L_{\text{Frequency}}(p_i, \hat{p}_i),\quad\text{otherwise,}
\end{cases}
\end{split}
\end{equation}

\noindent where $\hat{p}_i^{'}$ denotes the pixel in the attention-enhanced image. The variables \( \lambda \), \( \beta \) and \( \eta \) are the respective weights assigned to the loss components.
\begin{table*}[!htbp]
\centering
\caption{\textbf{Quantitative Comparison with Baselines on the Mip-NeRF 360 Dataset~\cite{barron2022mip}.} Each approach is rendered in four different resolutions (1/8, 1/4, 1/2, and the full resolution) after being trained at the lowest resolution (1/8). Our approach produces similar results at the 1/8 resolution and outperforms other models at 1/2, 1/4, and full resolutions.}
\resizebox{\textwidth}{!}{
\begin{tabular}{lccccc|ccccc|ccccc}
\hline
& \multicolumn{5}{c|}{SSIM $\uparrow$} & \multicolumn{5}{c|}{PSNR $\uparrow$} & \multicolumn{5}{c}{LPIPS $\downarrow$} \\
\cmidrule(lr){2-6} \cmidrule(lr){7-11} \cmidrule(lr){12-16}
& 1/8 Res. & 1/4 Res. & 1/2 Res. & Full Res. & Avg. & 1/8 Res. & 1/4 Res. & 1/2 Res. & Full Res. & Avg. & 1/8 Res. & 1/4 Res. & 1/2 Res. & Full Res. & Avg. \\
\hline
Instant-NGP & 0.748 & 0.645 & 0.620 & 0.690 & 0.676 & 26.85 & 24.90 & 24.15 & 24.40 & 25.08 & 0.238 & 0.373 & 0.452 & 0.466 & 0.382 \\
Mip-NeRF 360 & 0.858 & 0.730 & 0.665 & 0.700 & 0.738 & 29.24 & 25.31 & 24.08 & 24.17 & 25.70 & 0.125 & 0.263 & 0.368 & 0.431 & 0.297 \\
Zip-NeRF & 0.877 & 0.690 & 0.571 & 0.555 & 0.673 & 29.64 & 23.25 & 20.91 & 20.24 & 23.51 & 0.101 & 0.263 & 0.418 & 0.492 & 0.319 \\
\hline
3DGS & 0.882 & 0.735 & 0.616 & 0.622 & 0.714 & 29.25 & 23.44 & 20.80 & 19.52 & 23.25 & 0.105 & 0.242 & 0.396 & 0.483 & 0.307 \\
3DGS + EWA & 0.882 & 0.773 & 0.673 & 0.646 & 0.744 & 29.34 & 25.87 & 23.69 & 22.83 & 25.43 & 0.112 & 0.235 & 0.371 & 0.448 & 0.292 \\
Mip-Splatting & 0.881 & 0.799 & 0.713 & 0.723 & 0.779 & 29.29 & 26.99 & 26.03 & 25.59 & 26.98 & 0.108 & 0.221 & 0.315 & 0.403 & 0.262 \\
Ours & \textbf{0.888} & \textbf{0.813} & \textbf{0.749} & \textbf{0.771} & \textbf{0.805} & \textbf{29.85} & \textbf{27.26} & \textbf{26.40} & \textbf{26.28} & \textbf{27.45} & \textbf{0.098} & \textbf{0.208} & \textbf{0.301} & \textbf{0.384} & \textbf{0.248} \\
\hline
\end{tabular}
}
\label{tab:compare_mip_1}
\end{table*}
\section{Experiments}
\label{sec:experiment}
\subsection{Baselines, Datasets, Metrics and Implementation}
We selected Mip-Splatting~\cite{yu2024mip} and 3D-GS~\cite{kerbl20233d} as our primary baseline due to their established state-of-the-art performance in novel view synthesis. In our evaluation, we included several other prominent techniques, such as Mip-NeRF360~\cite{barron2022mip}, Instant-NGP~\cite{muller2022instant}, and Zip-NeRF~\cite{barron2023zip}. We also considered 3DGS + EWA~\cite{zwicker2001ewa} for further comparison.

We extensively evaluated multiple scenes from publicly available datasets, including a dataset that features a division of a large scene. Specifically, we assessed our method using seven scenes drawn from Mip-NeRF360~\cite{barron2022mip}, a Villa scene, Mill-19 dataset~\cite{turki2022mega} and two scenes from Tanks\&Temples~\cite{knapitsch2017tanks}. The Villa scene is a self-processed dataset. The evaluation metrics we report include Peak Signal-to-Noise Ratio (PSNR), Structural Similarity Index Measure (SSIM)~\cite{wang2004image}, and Learned Perceptual Image Patch Similarity (LPIPS)~\cite{zhang2018unreasonable}. 

Following 3DGS~\cite{kerbl20233d}, we trained our models for 30k iterations and applied patch attention every 50 iterations, and adjusted the scale matrix $S$ every 1,000 iterations up to 10,000 iterations. We set the kernel size as 0.05 and the loss weight as $\lambda = 0.2$.
\begin{figure}[t]
    \centering
    \includegraphics[width=\textwidth]{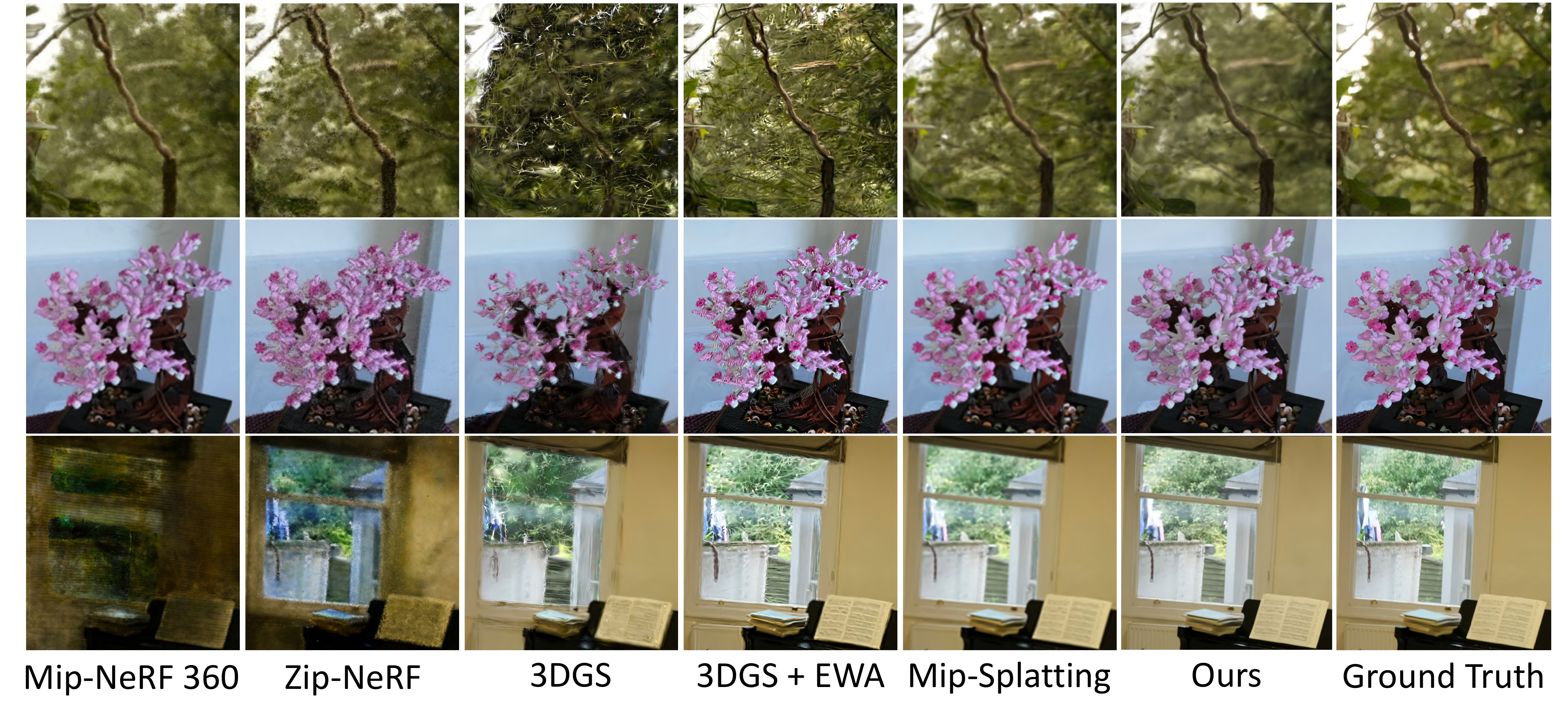}
    \caption{\textbf{Qualitative Comparison Results on the Mip-NeRF 360 Dataset~\cite{barron2022mip}.} These models were trained using images downsampled by a factor of eight and then rendered at full resolution to depict the quality of zooming in and close-ups. In contrast to previous approaches, our model achieves a higher level of accuracy and detail than other models and can render images that are almost identical to the ground truth.}
    \label{fig:mip-nerf-360}
\end{figure}
\begin{figure}[!htbp]
    \centering
    \includegraphics[width=\textwidth]{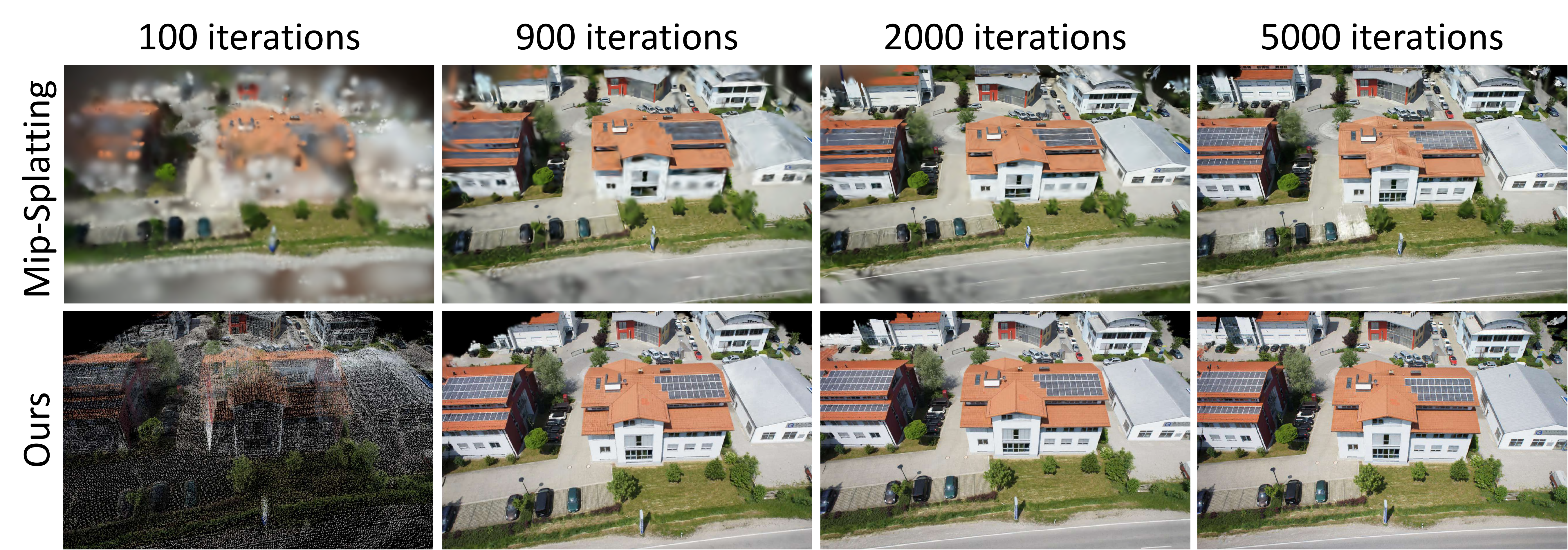}
    \caption{\textbf{Training Progression on the Villa Dataset.} We present the quality of the reconstructed villa scene at different training iterations. Compared to the SoTA Mip-Splatting~\cite{yu2024mip}, our method not only converges faster but also achieves better reconstruction quality with less noise.}
    \label{fig:villa}
\end{figure}
\subsection{Result Analysis}

%

\noindent\textbf{Comparison on the Mip-NeRF 360 Dataset}\ \ \ \
\noindent 
In our experiments, we trained models on data downsampled by a factor of eight and rendered images at multiple resolutions (1/8, 1/4, 1/2, and full). As shown in the Tab.~\ref{tab:compare_mip_1}, our method outperforms the baseline models in rendering images at both high and low resolutions. Fig.~\ref{fig:mip-nerf-360} shows our approach yields high-fidelity images without high-frequency artifacts, contrasting with Mip-NeRF 360~\cite{barron2022mip} and Zip-NeRF~\cite{barron2023zip}, which struggle at higher resolutions due to MLP limitations. Furthermore, traditional 3DGS~\cite{kerbl20233d} often suffers from degradation artifacts, and 3DGS + EWA~\cite{zwicker2001ewa} methods exhibit high-frequency artifacts. Our approach avoids these issues, more accurately represents the ground truth, and reduces blurred artifacts seen in Mip-Splatting~\cite{yu2024mip}.

\noindent\textbf{Comparison on the Villa Dataset}\ \ \ \
\noindent In the Villa Dataset experiment, we compared our model's training progression against Mip-Splatting~\cite{yu2024mip}, with both models trained at the original resolution. Results in Fig.~\ref{fig:villa} show the performance of both models across 100, 900, 2k, and 5k iterations. Our model demonstrated significant improvements by the 900th iteration, whereas scenes from Mip-Splatting~\cite{yu2024mip} remained blurry and of lower quality. Even after 5k iterations, Mip-Splatting failed to achieve the detail level of our model at 900 iterations, especially in fine features such as roofs, windows, and walls. \par
\begin{figure}[t]
    \centering
    \includegraphics[width=\textwidth]{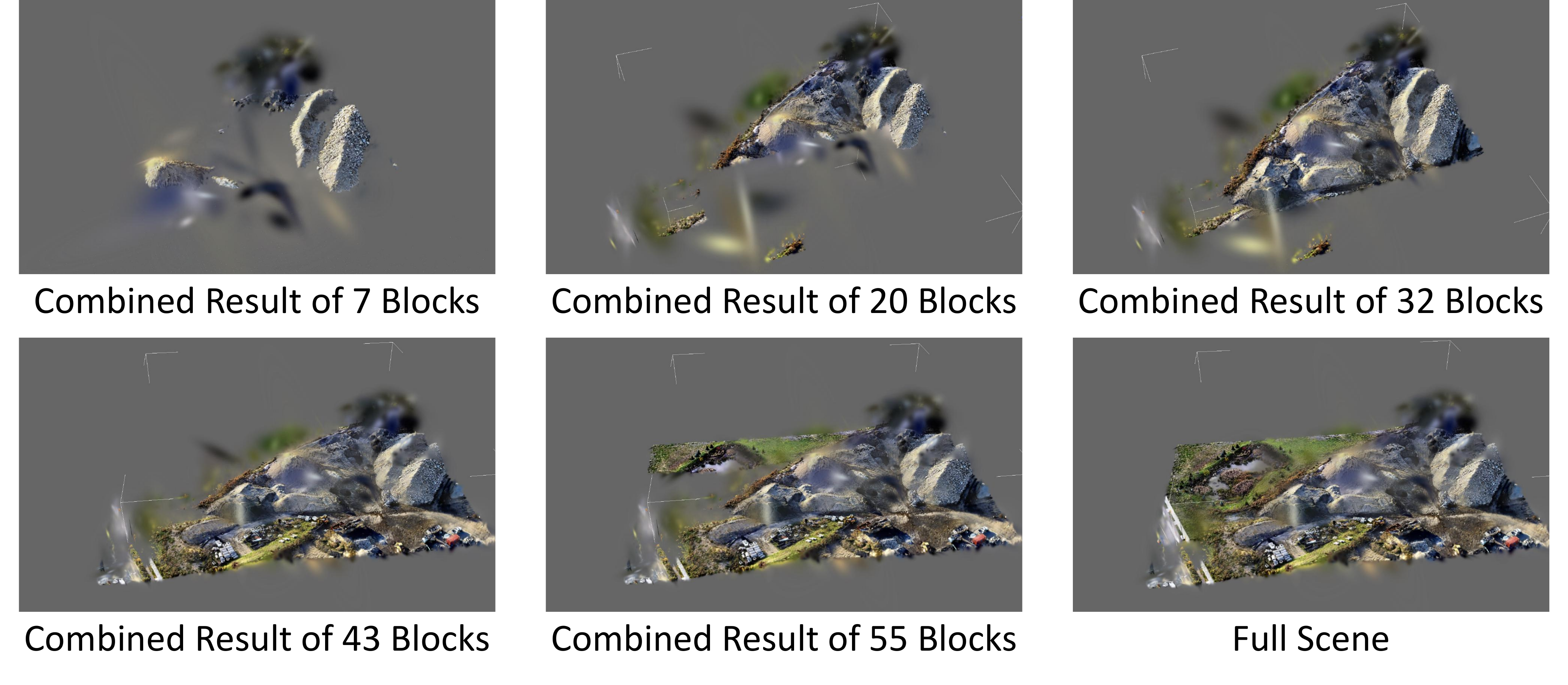}
    \caption{\textbf{Reconstructed Result on the Large Scene Dataset (Mill-19)~\cite{turki2022mega}.} We divide the large scene into individual blocks for separate reconstruction. Here, we display the recombined results of multiple blocks and the result of the full scene.}
    \label{fig:large_scene}
\end{figure}

\noindent\textbf{Evaluation on the Large Scene Dataset}\ \ \ \
\noindent In our study, we addressed the challenges of reconstructing large scenes like small towns or city-scale environments, which are unmanageable for traditional 3DGS-based~\cite{kerbl20233d} and NeRF-based~\cite{mildenhall2021nerf} models due to memory constraints and long optimization times. We used the Mill-19 Rubble scene~\cite{turki2022mega}, which had excessively noisy point clouds requiring reprocessing and selective image filtering. We subdivided the scene, which contained over 1,700 images, into 64 blocks. Each block was independently trained. This reduced memory demands and allowed efficient parallel training in just 20 minutes. As shown in Fig.~\ref{fig:large_scene}, our reconstruction results show the seamless reassembly of all blocks, which preserves the continuity of the large-scale scene. This method contrasts with previous models, which failed to directly reconstruct large scenes and compromised on reconstruction quality by randomly selecting a subset of images for training.

\begin{figure}[t]
    \centering
    \includegraphics[width=1\textwidth]{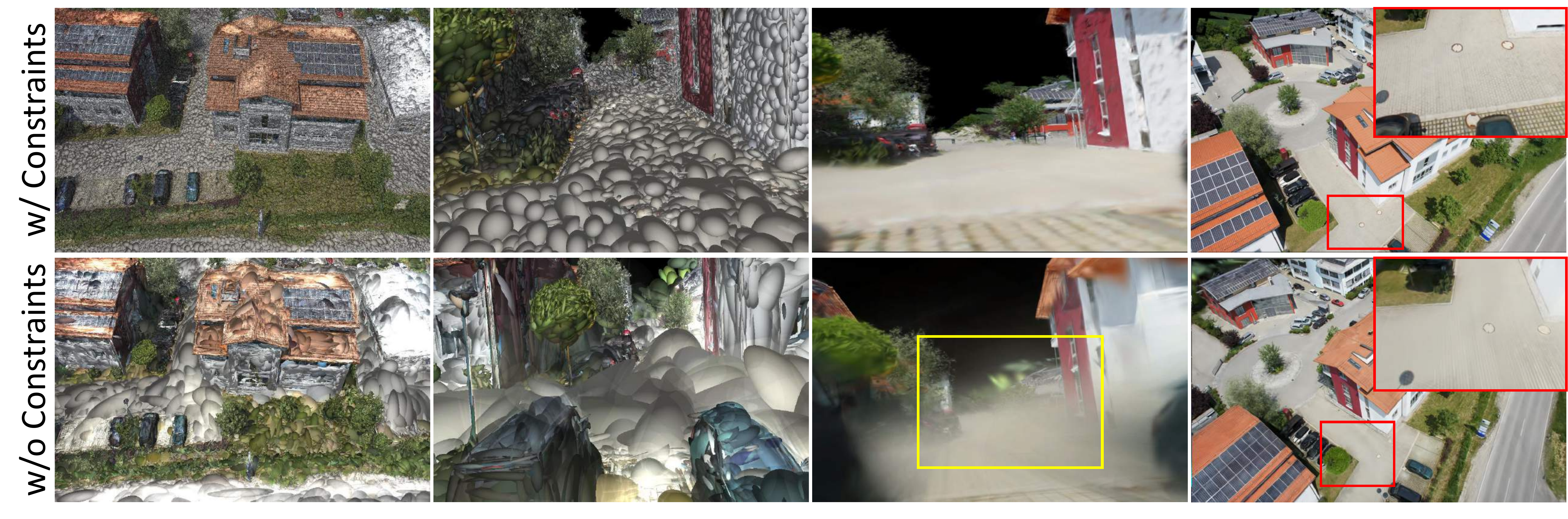}
    \caption{\textbf{Ablation of Gaussian Sphere Constraints Strategy.} We present an ablation study of our model trained on the Villa scene, comparing results with and without the application of the Gaussian Sphere Constraints Strategy. We present different viewing angles of the training Gaussians and the 3D scene. This strategy reduces the ``air walls'' problem.}
    \label{fig:ablation_gaussian}
\end{figure}
\subsection{Ablation Study}
\subsubsection{Patch Attention Enhancement}
Omitting the $L_{Frequency}$ and $L_{Edge}$ from the Patch Attention strategy leads to a significant reduction in rendering quality. To quantitatively evaluate this impact, we produced Table.~\ref{tab:ablation}, which compares performance metrics with and without this enhancement. The results clearly show that applying the Patch Attention strategy leads to consistent improvements across all metrics, significantly enhancing the model’s ability to generate sharper and more detailed renderings by focusing on edge information.
\begin{table*}[t]
\centering
\caption{\textbf{Ablation Study: Patch Attention Enhancement and Gaussian Sphere Constraints.} We present quantitative results for the Villa, Mip-NeRF 360 dataset~\cite{barron2022mip} and Tanks\&Temples dataset~\cite{knapitsch2017tanks}, trained for 30,000 iterations. Both scenes were downsampled by a factor of four and rendered at the same resolution.}
\begin{adjustbox}{width=\textwidth}
\begin{tabular}{lccc|ccc|ccc}
\hline
& \multicolumn{3}{c|}{Villa} & \multicolumn{3}{c|}{Mip-NeRF 360} & \multicolumn{3}{c}{Tanks\&Temples} \\
& SSIM $\uparrow$ & PSNR $\uparrow$ & LPIPS $\downarrow$ & SSIM $\uparrow$ & PSNR $\uparrow$ & LPIPS $\downarrow$ & SSIM $\uparrow$ & PSNR $\uparrow$ & LPIPS $\downarrow$\\
\hline
None & 0.855 & 25.30 & 0.202 & 0.813 & 26.33 & 0.188 & 0.810 & 22.17 & 0.191 \\
w/ Gaussian Constraints & 0.892 & 25.97 & 0.125 & 0.864 & 26.77 & 0.144 & 0.839 & 22.57 & 0.155 \\
w/ $L_{Frequency}$ & 0.888 & 25.85 & 0.131 & 0.874 & 26.98 & 0.128 & 0.856 & 23.29 & 0.141 \\
w/ $L_{Edge}$ & 0.873 & 25.67 & 0.138 & 0.861 & 26.91 & 0.131 & 0.843 & 23.15 & 0.143 \\
Full model & \textbf{0.893} & \textbf{26.43} & \textbf{0.121} & \textbf{0.881} & \textbf{27.44} & \textbf{0.111} & \textbf{0.863} & \textbf{23.77} & \textbf{0.133} \\
\hline
\end{tabular}
\end{adjustbox}
\label{tab:ablation}
\end{table*}

\subsubsection{Gaussian Sphere Constraints}
We assessed the importance of Gaussian Sphere Constraints by removing them from our model. As shown in Fig.~\ref{fig:ablation_gaussian}, models rendered without these constraints exhibit oversized Gaussian spheres, which result in information loss and reduce the overall quality of the renderings. In 3D scenes, these oversized spheres often create ``air walls'' in detail-heavy areas. Implementing Gaussian Sphere Constraints allows us to effectively control the growth and size of these spheres, enhancing detailed depiction within the scene. Fig.~\ref{fig:ablation_gaussian} clearly demonstrates the loss of detail in models rendered without this strategy. These images highlight how the constrained Gaussian spheres maintain finer details, leading to more precise and realistic renderings. Additionally, as indicated in Table.~\ref{tab:ablation}, the inclusion of Gaussian Sphere Constraints significantly improves performance metrics.

\subsubsection{Subdivision-Based Reconstruction of Large Scenes}
In this work, we are unable to conduct an ablation study for our ``Subdivision-Based Reconstruction of Large Scenes'' due to the inability of existing 3DGS-based methods~\cite{kerbl20233d,yu2024mip,lu2024scaffold,guedon2024sugar,jiang2024gs,huang2025structgs} to handle large-scale scenes without out-of-memory errors. Our subdivision strategy is essential for enabling 3DGS training on such scenes—a task infeasible for current methods without significant modifications. As this strategy is not a mere enhancement but a prerequisite for large-scale training, removing it for ablation would render the application of 3DGS on large scenes impossible.

\subsection{Limitation}
Our Patch Attention Enhancement increases memory demands and potentially causes out-of-memory errors during training, especially in complex scenes. Future model iterations could explore more efficient computational strategies to alleviate this issue. Additionally, assembling large scenes reveals overlaps at block boundaries, causing visual discontinuities. A future algorithm could remove Gaussian spheres at these boundaries to improve scene continuity.
\section{Conclusion}
\label{sec:conclusion}
In this paper, we present GaussianFocus, an enhanced model built upon traditional 3D Gaussian Splatting. It features three key innovations: Patch Attention Enhancement, Gaussian Constraints Strategy, and the subdivision of large-scale scenes into manageable blocks for individual training. These innovations aim to refine detail representation, enhance reconstruction quality and reduce the ``air walls'' problem. The approach of subdividing large scenes into manageable blocks overcomes the limitations inherent in traditional 3DGS-based methods, which struggle with extensive scenes. Experimental results demonstrate that GaussianFocus competes well with state-of-the-art methods at a single scale and excels across multiple scales, providing superior detail accuracy and reconstruction quality.

\begin{credits}
\subsubsection{\discintname}
The authors have no competing interests to declare that are relevant to the content of this article.
\end{credits}
%
%
%

\section{Acknowledgement}
This preprint has not undergone peer review (when applicable) or any post-submission improvements or corrections. The Version of Record of this contribution is published in Communications in Computer and Information Science (CCIS, volume 2756), and is available online at \href{https://doi.org/10.1007/978-981-95-4097-6_2}{https://doi.org/10.1007/978-981-95-4097-6\_2}.

\bibliographystyle{splncs04}
\bibliography{mybibliography}

\begin{thebibliography}{10}
\providecommand{\url}[1]{\texttt{#1}}
\providecommand{\urlprefix}{URL }
\providecommand{\doi}[1]{https://doi.org/#1}

\bibitem{barron2021mip}
Barron, J.T., Mildenhall, B., Tancik, M., Hedman, P., Martin-Brualla, R., Srinivasan, P.P.: Mip-nerf: A multiscale representation for anti-aliasing neural radiance fields. In: Proceedings of the IEEE/CVF international conference on computer vision. pp. 5855--5864 (2021)

\bibitem{barron2022mip}
Barron, J.T., Mildenhall, B., Verbin, D., Srinivasan, P.P., Hedman, P.: Mip-nerf 360: Unbounded anti-aliased neural radiance fields. In: Proceedings of the IEEE/CVF conference on computer vision and pattern recognition. pp. 5470--5479 (2022)

\bibitem{barron2023zip}
Barron, J.T., Mildenhall, B., Verbin, D., Srinivasan, P.P., Hedman, P.: Zip-nerf: Anti-aliased grid-based neural radiance fields. In: Proceedings of the IEEE/CVF International Conference on Computer Vision. pp. 19697--19705 (2023)

\bibitem{chen2022tensorf}
Chen, A., Xu, Z., Geiger, A., Yu, J., Su, H.: Tensorf: Tensorial radiance fields. In: European conference on computer vision. pp. 333--350. Springer (2022)

\bibitem{guedon2024sugar}
Gu{\'e}don, A., Lepetit, V.: Sugar: Surface-aligned gaussian splatting for efficient 3d mesh reconstruction and high-quality mesh rendering. In: Proceedings of the IEEE/CVF Conference on Computer Vision and Pattern Recognition. pp. 5354--5363 (2024)

\bibitem{huang2024efficient}
Huang, Z., Erfani, S.M., Lu, S., Gong, M.: Efficient neural implicit representation for 3d human reconstruction. Pattern Recognition p. 110758 (2024)

\bibitem{huang2025structgs}
Huang, Z., Xu, M., Perry, S.: Structgs: Adaptive spherical harmonics and rendering enhancements for superior 3d gaussian splatting. arXiv preprint arXiv:2503.06462  (2025)

\bibitem{jiang2024gs}
Jiang, Y., Li, J., Qin, H., Dai, Y., Liu, J., Zhang, G., Zhang, C., Yang, T.: Gs-sfs: Joint gaussian splatting and shape-from-silhouette for multiple human reconstruction in large-scale sports scenes. IEEE Transactions on Multimedia  (2024)

\bibitem{kerbl20233d}
Kerbl, B., Kopanas, G., Leimk{\"u}hler, T., Drettakis, G.: 3d gaussian splatting for real-time radiance field rendering. ACM Trans. Graph.  \textbf{42}(4),  139--1 (2023)

\bibitem{kerbl2024hierarchical}
Kerbl, B., Meuleman, A., Kopanas, G., Wimmer, M., Lanvin, A., Drettakis, G.: A hierarchical 3d gaussian representation for real-time rendering of very large datasets. ACM Transactions on Graphics (TOG)  \textbf{43}(4),  1--15 (2024)

\bibitem{knapitsch2017tanks}
Knapitsch, A., Park, J., Zhou, Q.Y., Koltun, V.: Tanks and temples: Benchmarking large-scale scene reconstruction. ACM Transactions on Graphics (ToG)  \textbf{36}(4),  1--13 (2017)

\bibitem{lassner2021pulsar}
Lassner, C., Zollhofer, M.: Pulsar: Efficient sphere-based neural rendering. In: Proceedings of the IEEE/CVF Conference on Computer Vision and Pattern Recognition. pp. 1440--1449 (2021)

\bibitem{lin2024vastgaussian}
Lin, J., Li, Z., Tang, X., Liu, J., Liu, S., Liu, J., Lu, Y., Wu, X., Xu, S., Yan, Y., et~al.: Vastgaussian: Vast 3d gaussians for large scene reconstruction. In: Proceedings of the IEEE/CVF Conference on Computer Vision and Pattern Recognition. pp. 5166--5175 (2024)

\bibitem{lu2024scaffold}
Lu, T., Yu, M., Xu, L., Xiangli, Y., Wang, L., Lin, D., Dai, B.: Scaffold-gs: Structured 3d gaussians for view-adaptive rendering. In: Proceedings of the IEEE/CVF Conference on Computer Vision and Pattern Recognition. pp. 20654--20664 (2024)

\bibitem{mildenhall2021nerf}
Mildenhall, B., Srinivasan, P.P., Tancik, M., Barron, J.T., Ramamoorthi, R., Ng, R.: Nerf: Representing scenes as neural radiance fields for view synthesis. Communications of the ACM  \textbf{65}(1),  99--106 (2021)

\bibitem{muller2022instant}
M{\"u}ller, T., Evans, A., Schied, C., Keller, A.: Instant neural graphics primitives with a multiresolution hash encoding. ACM transactions on graphics (TOG)  \textbf{41}(4),  1--15 (2022)

\bibitem{munkberg2022extracting}
Munkberg, J., Hasselgren, J., Shen, T., Gao, J., Chen, W., Evans, A., M{\"u}ller, T., Fidler, S.: Extracting triangular 3d models, materials, and lighting from images. In: Proceedings of the IEEE/CVF Conference on Computer Vision and Pattern Recognition. pp. 8280--8290 (2022)

\bibitem{reiser2021kilonerf}
Reiser, C., Peng, S., Liao, Y., Geiger, A.: Kilonerf: Speeding up neural radiance fields with thousands of tiny mlps. In: Proceedings of the IEEE/CVF international conference on computer vision. pp. 14335--14345 (2021)

\bibitem{schonberger2016structure}
Schonberger, J.L., Frahm, J.M.: Structure-from-motion revisited. In: Proceedings of the IEEE conference on computer vision and pattern recognition. pp. 4104--4113 (2016)

\bibitem{snavely2006photo}
Snavely, N., Seitz, S.M., Szeliski, R.: Photo tourism: exploring photo collections in 3d. In: ACM siggraph 2006 papers, pp. 835--846 (2006)

\bibitem{sun2022direct}
Sun, C., Sun, M., Chen, H.T.: Direct voxel grid optimization: Super-fast convergence for radiance fields reconstruction. In: Proceedings of the IEEE/CVF conference on computer vision and pattern recognition. pp. 5459--5469 (2022)

\bibitem{tancik2022block}
Tancik, M., Casser, V., Yan, X., Pradhan, S., Mildenhall, B., Srinivasan, P.P., Barron, J.T., Kretzschmar, H.: Block-nerf: Scalable large scene neural view synthesis. In: 2022 IEEE/CVF Conference on Computer Vision and Pattern Recognition. pp. 8248--8258 (2022)

\bibitem{turki2022mega}
Turki, H., Ramanan, D., Satyanarayanan, M.: Mega-nerf: Scalable construction of large-scale nerfs for virtual fly-throughs. In: 2022 IEEE/CVF Conference on Computer Vision and Pattern Recognition (CVPR). pp. 12912--12921. IEEE (2022)

\bibitem{wang2021neus}
Wang, P., Liu, L., Liu, Y., Theobalt, C., Komura, T., Wang, W.: Neus: Learning neural implicit surfaces by volume rendering for multi-view reconstruction. Advances in Neural Information Processing Systems  \textbf{34},  27171--27183 (2021)

\bibitem{wang2023neus2}
Wang, Y., Han, Q., Habermann, M., Daniilidis, K., Theobalt, C., Liu, L.: Neus2: Fast learning of neural implicit surfaces for multi-view reconstruction. In: IEEE/CVF International Conference on Computer Vision. pp. 3272--3283. IEEE (2023)

\bibitem{wang2004image}
Wang, Z., Bovik, A.C., Sheikh, H.R., Simoncelli, E.P.: Image quality assessment: from error visibility to structural similarity. IEEE transactions on image processing  \textbf{13}(4),  600--612 (2004)

\bibitem{weng2022humannerf}
Weng, C.Y., Curless, B., Srinivasan, P.P., Barron, J.T., Kemelmacher-Shlizerman, I.: Humannerf: Free-viewpoint rendering of moving people from monocular video. In: 2022 IEEE/CVF conference on Computer Vision and Pattern Recognition. pp. 16210--16220 (2022)

\bibitem{yifan2019differentiable}
Yifan, W., Serena, F., Wu, S., {\"O}ztireli, C., Sorkine-Hornung, O.: Differentiable surface splatting for point-based geometry processing. ACM Transactions on Graphics (TOG)  \textbf{38}(6),  1--14 (2019)

\bibitem{yu2021plenoctrees}
Yu, A., Li, R., Tancik, M., Li, H., Ng, R., Kanazawa, A.: Plenoctrees for real-time rendering of neural radiance fields. In: Proceedings of the IEEE/CVF International Conference on Computer Vision. pp. 5752--5761 (2021)

\bibitem{yu2024mip}
Yu, Z., Chen, A., Huang, B., Sattler, T., Geiger, A.: Mip-splatting: Alias-free 3d gaussian splatting. In: Proceedings of the IEEE/CVF Conference on Computer Vision and Pattern Recognition. pp. 19447--19456 (2024)

\bibitem{zhang2018unreasonable}
Zhang, R., Isola, P., Efros, A.A., Shechtman, E., Wang, O.: The unreasonable effectiveness of deep features as a perceptual metric. In: Proceedings of the IEEE conference on computer vision and pattern recognition. pp. 586--595 (2018)

\bibitem{zwicker2001ewa}
Zwicker, M., Pfister, H., Van~Baar, J., Gross, M.: Ewa volume splatting. In: Proceedings Visualization, 2001. VIS'01. pp. 29--538. IEEE (2001)

\end{thebibliography}

\end{document}